\documentclass[journal]{IEEEtran}

\IEEEoverridecommandlockouts
\usepackage{graphicx}
\def\BibTeX{{\rm B\kern-.05em{\sc i\kern-.025em b}\kern-.08em
    T\kern-.1667em\lower.7ex\hbox{E}\kern-.125emX}}
\usepackage[font=small,labelfont=bf,tableposition=top]{caption}

\usepackage{blindtext}
\usepackage{caption}

\let\oldtwocolumn\twocolumn
\renewcommand\twocolumn[1][]{%
    \oldtwocolumn[{#1}{
    \begin{center}
    \vskip-3ex
        \centering
        \includegraphics[scale=0.4]{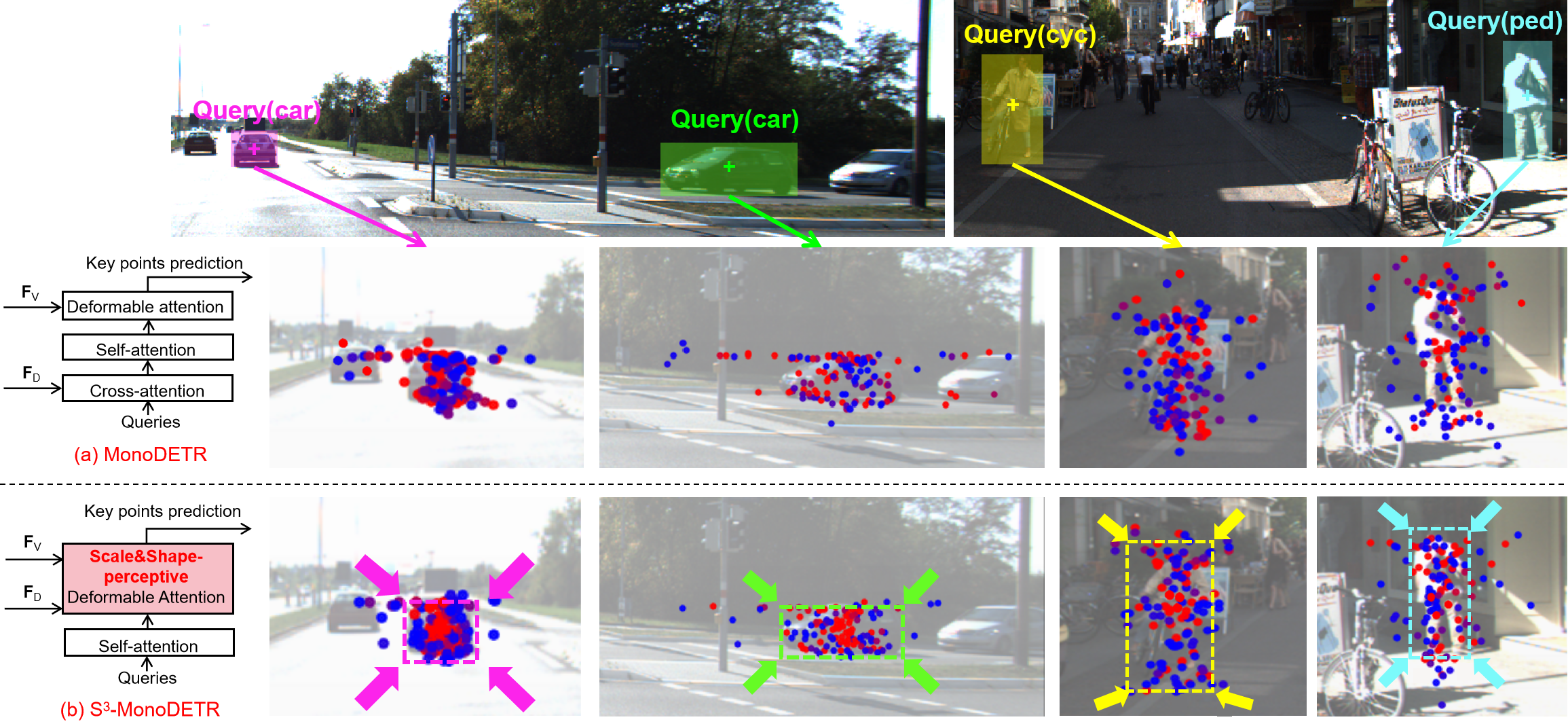}
        \captionof{figure}{Four examples to visualize the predicted key points (cycle) on object queries (cross) by (a) MonoDETR and (b) S$^3$-MonoDETR, where the change in key point color from red to blue indicates a shift in attention weight from high to low. It is clear that S$^3$-DA could generate key points with higher quality by using the shape$\&$scale-perceptive mechanism.}
        \label{intro}
    \end{center}
    }]
}

\usepackage{times}
\usepackage{epsfig}
\usepackage{graphicx}
\usepackage{amsmath}
\usepackage{amssymb}
\usepackage{color}
\usepackage{booktabs}
\usepackage{bbm}
\usepackage{bm}
\usepackage{multirow}
\usepackage{indentfirst}
\usepackage{hhline}
\usepackage{array}
\usepackage{arydshln}
\newcolumntype{C}[1]{>{\centering}p{#1}}
\usepackage{makecell}
\usepackage{pifont}
\usepackage{caption}
\usepackage{graphicx}
\usepackage{float}
\usepackage{subcaption}
\usepackage{mathrsfs}
\usepackage[table]{xcolor}
\definecolor{rblue}{rgb}{0,0.5,1}
\usepackage[breaklinks,colorlinks]{hyperref}
\hypersetup{colorlinks, citecolor=rblue}

\begin{document}
\title{S$^3$-MonoDETR:\\Supervised Shape$\&$Scale-perceptive Deformable Transformer for Monocular 3D Object Detection}

\author{Xuan He$^{1}$, Jin Yuan$^{2}$, Kailun Yang$^{3}$, Zhenchao Zeng$^{1}$, and Zhiyong Li$^{2,3}$%
\thanks{$^{1}$X. He and Z. Zeng are with the Hunan Vanguard Group Corporation Limited, Changsha 410100, PR China.}%
\thanks{$^{2}$J. Yuan and Z. Li are with the College of Computer Science and Electronic Engineering, Hunan University, Changsha 410082, China.}%
\thanks{$^{3}$K. Yang and Z. Li are with the School of Robotics, Hunan University, Changsha 410012, China.}%
}

\maketitle

\begin{abstract}
Recently, transformer-based methods have shown exceptional performance in monocular 3D object detection, which can predict 3D attributes from a single 2D image. These methods typically use visual and depth representations to generate query points on objects, whose quality plays a decisive role in the detection accuracy. However, current unsupervised attention mechanisms without any geometry appearance awareness in transformers are susceptible to producing noisy features for query points, which severely limits the network performance and also makes the model have a poor ability to detect multi-category objects in a single training process. To tackle this problem, this paper proposes a novel ``Supervised Shape$\&$Scale-perceptive Deformable Attention'' (S$^3$-DA) module for monocular 3D object detection. Concretely, S$^3$-DA utilizes visual and depth features to generate diverse local features with various shapes and scales and predict the corresponding matching distribution simultaneously to impose valuable shape$\&$scale perception for each query. Benefiting from this, S$^3$-DA effectively estimates receptive fields for query points belonging to any category, enabling them to generate robust query features. Besides, we propose a Multi-classification-based Shape$\&$Scale Matching (MSM) loss to supervise the above process. Extensive experiments on KITTI and Waymo Open datasets demonstrate that S$^3$-DA significantly improves the detection accuracy, yielding state-of-the-art performance of single-category and multi-category 3D object detection in a single training process compared to the existing approaches. The source code will be made publicly available at \url{https://github.com/mikasa3lili/S3-MonoDETR}.
\end{abstract}


\IEEEpeerreviewmaketitle

\section{Introduction}
\IEEEPARstart{T}{he} recent advancements in 3D object detection have greatly increased its applicability in autonomous driving~\cite{zhu2022vpfnet, liu2023multi, chen2023shapeaware_3D, gao2022camrl} and indoor robotics~\cite{song2015sun, dai2017scannet, wang2023net, xie2023farp}.
Although 3D object detection methods based on LiDAR points~\cite{zhou2018voxelnet, sheng2021improving, xu2022behind} or binocular images~\cite{li2019stereo, zhou2020salient, liu2021yolostereo3d} have achieved excellent performance, they come with high hardware costs. Thus, monocular 3D object detection methods~\cite{roddick2018orthographic, weng2019monocular, liu2020smoke}, which predict 3D attributes of objects from a single 2D image with significantly reduced computation and equipment costs, making it an area of increasing research interest.\\
\indent Most current monocular 3D detection methods are based on traditional 2D object detectors~\cite{lin2017focal, ren2015faster, tian2019fcos}.
Inspired by the great success of transformers in 2D object detection~\cite{chen2021pix2seq, zheng2020end, zhu2020deformable}, there are methods~\cite{huang2022monodtr, wu2022dst3d, wang2021pnp} endeavoring to introduce transformers into monocular 3D object detection.
For example, MonoDETR~\cite{zhang2022monodetr} employs several attention operations on visual and depth maps (see Fig.~\ref{intro}(a)) to sample hundreds of key points for each input query point, whose approach mainly originates from the Deformable DETR~\cite{zhu2020deformable}.
All the sampled key points of a query could represent its respective field, and the visual features of these key points are then integrated to predict the 3D attributes of this query.
Thus, the sampling accuracy of key points is crucial.

However, the original deformable attention mechanism exists a serious noisy point sampling problem, which arises from the fact that deformable attention only focuses on exploring relative key points for a query.
Yet, it ignores estimating the scope of the receptive field of the query, resulting in noisy key points outside the object, which is especially fatal for those small objects (for example, the pink car depicted in Fig.~\ref{intro}).
Moreover, we observe that the existing methods \cite{zhang2021objects, lu2021geometry, zhang2022monodetr} always train separate models with carefully tuned hyper-parameters for different categories of objects, due to there are significant differences in geometric appearance between them.
Undeniably, this can achieve better detection performance, but it is notoriously inefficient and impractical in safety-critical autonomous driving applications. 

\indent To alleviate these problems, we propose a Supervised Shape$\&$Scale-perceptive Deformable Transformer (S$^3$-MonoDETR) for monocular 3D object detection.
Different from MonoDETR~\cite{zhang2022monodetr}, S$^3$-MonoDETR reduces the depth cross-attention layer and introduces a novel Supervised Shape$\&$Scale-perceptive Deformable Attention (S$^3$-DA) layer to carry out precise key points prediction for each query (see Fig.~\ref{intro}(b)). 
Specifically, S$^3$-DA begins by generating several masks with varying shapes and scales to extract diverse local features from the input visual map for the input queries. Additionally, the model predicts the shape$\&$scale matching distribution of the queries guided by the fused depth and visual information, which is obtained from a novel query-level learnable fusion strategy.
With the combination of diverse local features and matching distribution, S$^3$-DA leverages a lightweight adaptive learning layer to generate a shape$\&$scale-aware filter to augment the queries, thereby providing valuable geometric appearances for their key point prediction.

To further enhance the effectiveness of S$^3$-DA, we introduce a Multi-classification-based Shape$\&$Scale Matching (MSM) loss to directly supervise the matching distribution learning without incurring extra labeling costs.
The proposed MSM loss helps S$^3$-DA carry out a scope-perception visual cues assignment for the input queries, thus enabling them to own accurate receptive fields with higher-quality features for 3D attribute prediction.

Based on the above designs, our model achieves better detection results for objects of various geometric appearances, which also enables it to outperform other models when performing the challenging multi-category joint training task.
An extensive set of experiments conducted on KITTI~\cite{geiger2012we} and Waymo Open~\cite{ettinger2021large} datasets verifies the effectiveness of the proposed method.

At a glance, the main contributions of this work are summarized as follows:
\begin{enumerate}
	\item We propose a Supervised Shape$\&$Scale-perceptive Deformable Attention (S$^3$-DA) with an MSM loss to improve the quality of queries in the transformer. As compared to deformable attention, S$^3$-DA could well estimate the receptive field of a query to support accurate key point generation, yielding high-quality query features for 3D attribute prediction.
    \item This is the first monocular 3D object detector dedicated to detecting objects from different categories with various geometric appearances in a single training process.
	\item Comprehensive experimental results on KITTI and Waymo Open datasets demonstrate the leading performance of our proposed approach compared to state-of-the-art approaches in moderate and hard subsets. Furthermore, the near real-time inference speed indicates the high applicability of our method in autonomous driving applications.
\end{enumerate}

\begin{figure*}[t]
	\centering
	\includegraphics[scale=0.42]{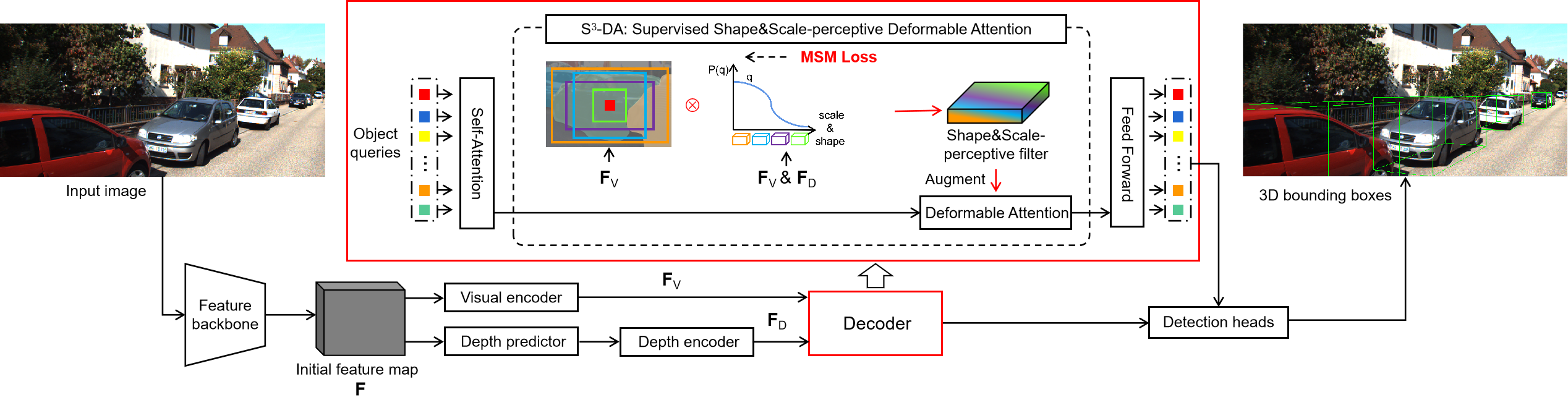}
	\caption{The architecture of S$^3$-MonoDETR, where S$^3$-DA with an MSM loss is introduced to generate a shape$\&$scale-aware filter to help the object queries yield more robust query features.}
	\label{pipeline}
\end{figure*}

\section{Related Work}
\textbf{Monocular 3D object detection via 2D detectors.}
The existing monocular 3D object detection approaches typically modify 2D object detectors to enable 3D attributes detection from a single image~\cite{chen2016monocular, chen20153d, simonelli2019disentangling, brazil2019m3d, shi2021geometry}.
For example,  OFTNet~\cite{roddick2018orthographic} maps the image-based features into an orthographic 3D space by introducing the orthographic feature transform.
M3DSSD~\cite{luo2021m3dssd} alleviates the feature mismatching problem by using a two-step feature alignment.
Recent methods~\cite{mousavian20173d, li2020rtm3d, lian2022monojsg} have explored effective but complex geometric priors to improve detection accuracy.
MonoPair~\cite{chen2020monopair} introduces the uncertainty theory and explores spatial relationships between pairs of objects.
GUPNet~\cite{lu2021geometry} introduces geometry-guided depth uncertainty and utilizes a hierarchical learning strategy to enhance the training process.
MonoFlex~\cite{zhang2021objects} proposes a flexible framework to explicitly decouple the truncated objects and consists of multiple approaches for the crucial depth estimation.
Despite their successes, these image-based approaches often face performance limitations due to the monocular nature of the input image.
Therefore, researchers have started incorporating external resources such as depth maps~\cite{ma2020rethinking, park2021pseudo, ding2020learning, ma2019accurate} and LiDAR~\cite{huang2022monodtr, reading2021categorical, chen2021monorun} to boost the detection performance.
However, these additions come with extra costs in terms of data collection and computation.

\textbf{Monocular 3D object detection via transformers.}
The transformer-based methods for 2D object detection have presented tremendous success~\cite{zhu2020deformable, gao2021fast, carion2020end}.
Inspired by them, recent researchers have increasingly turned their attention to transformers for monocular 3D object detection.
These approaches can eliminate the need for cumbersome post-processing steps such as Non-Maximum Suppression (NMS).
For example, MonoDTR~\cite{huang2022monodtr} proposes to learn depth-aware features with the help of LiDAR point clouds and introduces a depth-aware transformer module to globally integrate context- and depth-aware features.
Without any extra data, DST3D~\cite{wu2022dst3d} proposes to combine the Swin Transformer~\cite{liu2021swin} with deep layer aggregation~\cite{yu2018deep} to achieve 3D object detection.
MonoDETR~\cite{zhang2022monodetr} designs a depth-guided decoder, in which a depth cross-attention layer is used to learn global-aware depth features, and a deformable attention layer is introduced to aggregate local visual features for the input object queries. Through the innovative depth-guided decoder, MonoDETR achieves excellent performance.
However, in MonoDETR, the location and feature extraction of queries are entirely dependent on several unsupervised attention layers, which have limitations when it comes to detecting objects of large-span sizes, as well as different categories with significant appearance differences.

To alleviate the above difficulties, this paper focuses on improving the quality of query points in MonoDETR by imposing a supervised shape$\&$scale perception.
We introduce a novel approach called Supervised Shape$\&$Scale-perceptive Deformable Attention (S$^3$-DA), which replaces the deformable attention in the decoder of MonoDETR. The S$^3$-DA could flexibly predict accurate scopes of the receptive fields of queries, thereby aggregating high-quality visual features for queries to carry out robust 3D attribute prediction for various appearance objects.

\textbf{Feature fusion strategies.}
Existing feature fusion strategies \cite{he2015spatial, zhao2017pyramid, hu2018squeeze}
aim to fuse feature maps with different scales or semantic information to facilitate the subsequent tasks. The most primitive and simplest operation is to directly concatenate or add different feature maps.
To achieve more effective and robust feature fusion, FPN~\cite{lin2017feature} proposes to add horizontal links in the feature pyramids at different levels.
DANet~\cite{fu2019dual} introduces spatial and channel attention mechanisms to select key regions and key channels.
GC-Net~\cite{ni2020gc} utilizes a global average pooling method to capture global context information and fuse them with local features when combining multiple feature maps.
PANet~\cite{liu2018path} extracts features by constructing bottom-up and top-down paths on multiple feature maps, then fuses them through path aggregation. Although effective, the above approaches are dedicated to studying the fusion of the entire map level which needs extensive parameter learning.
In our work, to achieve a better perception of queries with different geometric appearances, we intend to fuse the depth and visual feature maps in a finer-grained and lightweight manner to provide thorough guidance for the shape$\&$scale matching distribution prediction of queries.
To this end, a novel query-level learnable feature fusion mechanism is proposed.

In general, our method is proposed with the aim of improving the quality of queries with several carefully designed components. Unlike the previous work SSD-MonoDETR~\cite{he2023SSD-MonoDTR}, which extracts local visual features by setting square regions of different scales, and uses the depth information to guide the prediction of scales for queries, the proposed method can extract more diverse local features of different shapes and scales, and proposes a query-level depth$\&$visual cues learnable fusion technology as well as an MCM loss to supervise the estimating of shapes and scales, providing the model with the ability to detect multiple categories of objects simultaneously in a single training process, which has more practical value.

\section{Methodology}

\subsection{Architecture Overview}
Fig.~\ref{pipeline} shows the architecture of the proposed Supervised Shape$\&$Scale-perceptive Deformable Transformer for monocular 3D object detection (S$^3$-MonoDETR). Specifically, a feature backbone is first employed to extract the initial feature $\mathbf{F}$ for an input image.
Based on $\mathbf{F}$, we utilize a visual encoder and a depth predictor with a depth encoder to generate global visual and depth features respectively. These features are then fed to the decoder to produce high-quality queries for 3D attribute prediction through several detection heads. 
Compared to MonoDETR~\cite{zhang2022monodetr}, S$^3$-MonoDETR proposes a novel Supervised Shape$\&$Scale-perceptive Deformable Attention (S$^3$-DA) module to merge valuable geometric appearance-aware information into each object query, thereby yielding more precise 3D attributes across all categories of objects. What is more, a Multi-classification-based Scale Matching (MSM) loss is introduced in S$^3$-DA to help it learn distinctive and credible shape$\&$scale perceptions for object queries. 

\begin{figure*}[ht]
	\centering
	\includegraphics[scale=0.45]{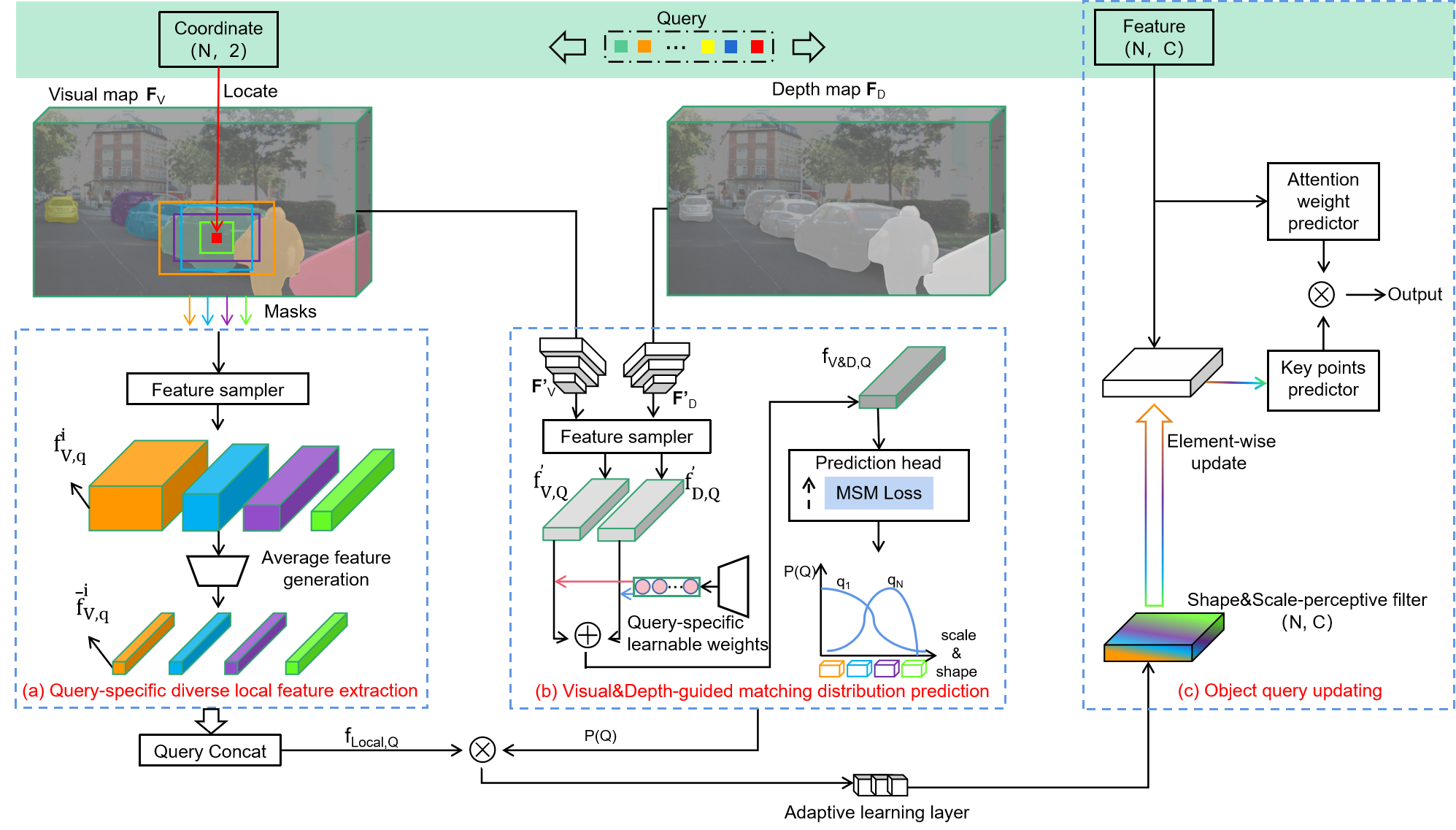}
	\caption{The specific design of S$^3$-DA, which is composed of three steps: (a) query-specific diverse local feature extraction, (b) visual$\&$depth-guided matching distribution prediction, and (c) object query updating.}
	\label{S$^3$-DA}
\end{figure*}

\subsection{Supervised Shape$\&$Scale-perceptive Deformable Attention Module}\label{DVMG}
S$^3$-DA belongs to the decoder in S$^3$-MonoDETR. It receives the visual embedding $\mathbf{F}_V$, the depth embedding $\mathbf{F}_D$, and a set of learnable queries $\mathbb{Q}$ from the last self-attention layer. Specifically, S$^3$-DA first extracts diverse local visual features with multi-shape and multi-scale cues for each object query in $\mathbb{Q}$ (see Sec.~\ref{QMF}) and predicts a visual$\&$depth-guided matching distribution at different shapes and scales (see Sec.~\ref{DMS}). Then, S$^3$-DA integrates diverse local features with matching distribution and utilizes an adaptive feature learning layer to form a shape$\&$scale-aware filter, which is used for object query updating (see Sec.~\ref{update}).

\subsubsection{Query-specific Diverse Local Feature Extraction}\label{QMF}
As shown in Fig.~\ref{S$^3$-DA}, given the global visual representation $\mathbf{F}_V \in \mathbbm{R}^{\frac{H}{16} \times \frac{W}{16} \times C}$, where $H$, $W$, and $C$ are the height, width, and the channel number of an input image.
This step aims to generate diverse local visual features for query $q \in \mathbb{Q}$.
Here, each query $q$ is randomly initialized with a feature embedding $f_{q} \in \mathbbm{R}^C$, and a position coordinate $(q_x, q_y)$.
Our approach first takes the query $q$ as the center to generate $I$ masks $\{M_{q,i}\}_{i=1}^I$ with the size of $r_iw_i \times w_i$, where $r_i$, $w_i$ are the aspect ratio and width of the $i{-}th$ mask, and are used to represent the shape and scale of a mask in the following parts.
Then, the local feature embedding $f_{V,q}^i \in \mathbbm{R}^{r_iw_i \times w_i \times C}$ is generated as follows:
\begin{equation}
	f_{V,q}^i = \mathcal{S}(\mathbf{F}_V, M_{q,i}),
	\label{sampler}
\end{equation}
where $\mathcal{S}$ is a feature sampler based on bilinear interpolation.
For total $(r_i*w_i+1) * (w_i+1)$ elements within $M_{q,i}$, $\mathcal{S}_V$ extracts the corresponding feature embedding from $\mathbf{F}_V$ according to their coordinates. This generates a local visual embedding $f_{V,q}^i$ to capture the visual representations for $q$ at the scope of $r_iw_i \times w_i$.
Finally, we integrate all the local embeddings in $f_{V,q}^i$ to calculate an average local feature $\bar{f}_{V,q}^i \in \mathbbm{R}^{C}$ as follows:
\begin{equation}
	\bar{f}_{V,q}^i = \frac{\sum_{(r_i*w_i+1) * (w_i+1)} f_{V,q}^i}{(r_i*w_i+1) * (w_i+1)}.
	\label{f}
\end{equation}
\indent For each query $q$, our approach utilizes $I$ masks to generate $I$ local feature embeddings $\{\bar{f}_{V,q}^i\}_{i=1}^I$, which could well capture diverse local feature information around $q$. Finally, we concatenate the local feature embedding of all the $q$ in $\mathbb{Q}$ to obtain a complete diverse local feature map $f_{Local, \mathbb{Q}} \in \mathbbm{R}^{N \times I \times C}$, which can be seen in Fig.~\ref{S$^3$-DA}.

\subsubsection{Visual$\&$Depth-guided Matching Distribution Prediction} \label{DMS}
Intuitively, the visual cues could reflect the geometric shape of objects, and the depth cues represent their distance to the camera and thus could reflect their scale. On this basis, a novel fusion mechanism is used to generate the fused visual and depth features, followed by a prediction head to generate the shape$\&$scale matching distribution, in which an MCM loss is designed to supervise this process.\\
\indent \textbf{Query-level Fusion of Visual and Depth Features.} 
As aforementioned, to carry out a finer-grained fusion between the depth and visual feature maps, so that to achieve thorough guidance for the shape$\&$scale prediction of queries with various appearances, we design a learnable query-level feature fusion mechanism to adaptively fuse visual and depth features according to their distinctive influence on different queries. This mechanism could help the fusion process become detailed query-level rather than the traditional coarse whole map-level. It also greatly reduces the number of parameters that need to be learned, resulting in a more lightweight model.\\
\indent Technically, as shown in Fig.~\ref{S$^3$-DA}(b), given the visual representation $\mathbf{F}_V \in \mathbbm{R}^{\frac{H}{16} \times \frac{W}{16} \times C}$ and depth representation $\mathbf{F}_D \in \mathbbm{R}^{\frac{H}{16} \times \frac{W}{16} \times C}$, we first use two convolutional layers on them to extract more compact features $\mathbf{F}'_V \in \mathbbm{R}^{\frac{H}{64} \times \frac{W}{64} \times C}$ and $\mathbf{F}'_D \in \mathbbm{R}^{\frac{H}{64} \times \frac{W}{64} \times C}$. Then, we employ a feature sampler $\mathcal{S}$ to extract the feature embedding $f'_{V, \mathbb{Q}} \in \mathbbm{R}^{N \times C}$, $f'_{D, \mathbb{Q}} \in \mathbbm{R}^{N \times C}$ from $\mathbf{F}'_{V}$, $\mathbf{F}'_{D}$ for all queries $\mathbb{Q}$:
\begin{equation}
	f'_{V, \mathbb{Q}} = \mathcal{S}(\mathbf{F}'_{V}, \mathbb{Q}),
	\label{sampler-v}
\end{equation}
\begin{equation}
	f'_{D, \mathbb{Q}} = \mathcal{S}(\mathbf{F}'_{D}, \mathbb{Q}).
	\label{sampler-d}
\end{equation}

\indent On this basis, we devise a learnable weight layer $\mathcal{L}$, which is implemented via a parameter generator with gradient feedback, and could generate a set of query-specific weights $\mathbb{W} \in \mathbbm{R}^{N \times 1}$ to adaptively learn the fusion proportion of visual and depth representations for each query. This process could be expressed by:
\begin{equation}
	\mathbb{W} = \mathcal{L}(N, 1),
	\label{bias}
\end{equation}
we set $\mathcal{L}$ with the initialized constant parameter $0.5$, which means the initial proportion of visual and depth information is equal.
Based on this, we multiply the corresponding weights with visual features and depth features, and add them up to obtain the query-level fusion feature $f_{V\&D,Q} \in \mathbbm{R}^{N \times C}$:
\begin{equation}
	f_{V\&D,\mathbb{Q}} = \mathbb{W}*f'_{V,\mathbb{Q}} + (\mathbb{E} - \mathbb{W})*f'_{D,\mathbb{Q}},
	\label{ff}
\end{equation}
where $\mathbb{E} \in \mathbbm{R}^{N \times 1}$ is an identity matrix. Finally, we design a prediction head to map $f_{V\&D,\mathbb{Q}}$ to a shape$\&$scale matching distribution $P(\mathbb{Q})$, which is formulated as follows:
\begin{equation}
	P(\mathbb{Q}) = \sigma(f_{V\&D,\mathbb{Q}}),
	\label{predictor}
\end{equation}
where $\sigma$ is a project function implemented via a linear layer. $P(\mathbb{Q})$ is a $N \times I$ vector to represent the $I$-category matching distribution. \\

\begin{figure}[t]
	\centering
	\includegraphics[scale=0.57]{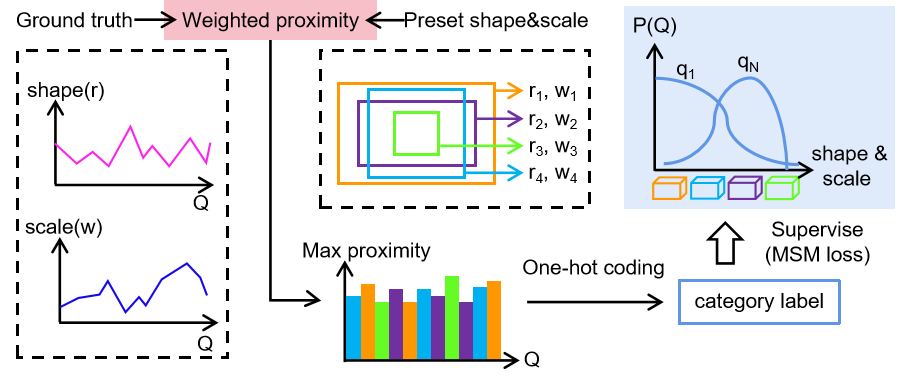}
	\caption{The process of category label generation for the MCM loss.}
	\label{myloss1}
\end{figure}

\textbf{Multi-classification-based \ Shape$\&$Scale \ Matching \ loss.}
\indent To make the generated shape$\&$scale matching distribution $P(\mathbb{Q})$ be consistent with the ground-truth values of object shapes and scales, a straightforward solution is to use an L1 loss between the true shape$\&$scale $\hat{S}(\mathbb{Q})$ and the predicted shape$\&$scale $S'(\mathbb{Q})$, and $S'(\mathbb{Q})$ could be approximately expressed by the sum of the corresponding multiplication between $P(\mathbb{Q})$ and preset shape$\&$scale $\mathbb{S} = \{[r_i, w_i] \}_{i=0}^I$:
\begin{equation}
	S'(\mathbb{Q}) = \sum_{i=0}^I P(\mathbb{Q}) * \mathbb{S}.
\label{predicted scale}
\end{equation}
\indent However, the preset shape$\&$scale could only be rough and coarse, and the true values of objects are continuous and endless, thus the meticulous L1  regression loss will be very difficult to fit.
To this end, we propose a Multi-classification-based Shape$\&$Scale Matching (MSM) loss to directly supervise the prediction of matching distribution $P(\mathbb{Q})$.
As shown in Fig.~\ref{myloss1}, the MSM loss first maps all the true shape$\&$scale $\hat{S}$ into $I$ categories according to their weighted proximity to the preset shape$\&$scale $\mathbb{S}$, where the weighted proximity operation aims to give different weights to the shape and scale when calculating the difference between $\hat{S}$ and $\mathbb{S}$, since the shape $w$ often has larger magnitude compared to the scale $r$. Finally, a one-hot coding is followed to generate the shape$\&$scale category label $\hat{C}$.
The process of category label generation could be expressed as:
 \begin{equation}
	\hat{C} = \mathbbm{O}(Index(\mathcal{W}(\hat{S}[r, w], \mathbb{S}[r, w]))),
\label{category_label_1}
\end{equation}
where $\mathbbm{O}$ is the one-hot coding operation and $Index()$ will back the index $i$ for the values in parentheses. $\mathcal{W}$ calculates the minimum weighted distance between the true shape$\&$scale $\hat{S}[r_i, w_i]$ and preset shape$\&$scale $\mathbb{S}[r_i, w_i]$, which could be expressed as:
 \begin{equation}
	\mathcal{W} = min(\{W_1(\hat{S}[r_i]-\mathbb{S}[r_i]) + W_2(\hat{S}[w_i]-\mathbb{S}[w_i])\}_{i=0}^I),
\label{category_label_2}
\end{equation}
where $W_1$ and $W_2$ are the balancing weights for shape and scale, respectively. Due to the scales and shapes of objects being most concentrated on the medium level, we utilize a multi-classification focal loss (FL) \cite{lin2017focal} to balance all scales and shapes for the supervision of $P(\mathbb{Q})$:
\begin{equation}
	L_{MSM} = \frac{1}{N}FL(\hat{C}(\mathbb{Q}), P(\mathbb{Q})).
\label{MSM}
\end{equation}

\subsubsection{Object Query Updating} \label{update}
Given an initial query, the query updating aims to use the query features to search for hundreds of key points around the query and integrate them with attention weights to represent the new query features. As mentioned earlier, while MonoDETR utilizes depth and visual features for key point prediction, it still suffers from the presence of numerous noisy key points.
To address this issue, as depicted in Fig.~\ref{S$^3$-DA}(c), our query updating strategy leverages diverse local visual features and shape$\&$scale matching distribution to adaptively learn a shape$\&$scale-aware filter for the augmentation of the original query features, which facilitates the identification of more accurate key points.
Specifically, our approach first combines the diverse local features with the matching distribution of all queries and utilizes an adaptive learning layer that incorporates a lightweight convolutional network with an activation layer to generate the shape$\&$scale-aware filter.
Subsequently, we perform an element-wise multiplication between the filter and the initial query feature to produce shape$\&$scale-perceptive query features.
Finally, the query feature is updated by incorporating these key point features along with the attention weights.
By leveraging the shape$\&$scale-aware filter, our approach is capable of accurately predicting key points and avoiding the addition of noisy information, thereby facilitating the generation of high-quality query features for improved 3D attribute prediction.

\subsection{Training Loss} \label{loss}
S$^3$-MonoDETR is an end-to-end network, and all the components are jointly trained according to a composite loss function, which consists of a query loss in detection heads to regression seven parts for each positive query, and the proposed WSM loss is equipped on the S$^3$-DA module to supervise the shape\&scale matching. Specifically, the query loss $L_{query}$ utilizes focal loss~\cite{lin2017focal} to estimate the object classes, L1 loss to estimate the 2D sizes $(l, r, t, b)$ and projected 3D center $(x_{3d}, y_{3d})$, GIoU loss~\cite{rezatofighi2019generalized} for 2D box IoU, Laplacian aleatoric uncertainty loss~\cite{chen2020monopair} for depth value, and follows MonoDLE~\cite{ma2021delving} to predict 3D sizes $(h_{3d}, w_{3d}, l_{3d})$ and orientation angle $\alpha$. Thus, $L_{query}$ is expressed as:
\begin{equation}
\begin{aligned}
	L_{query} &=  \lambda_1 L_{class} + \lambda_2 L_{2dsize} + \lambda_3 L_{xy3d} + \lambda_4 L_{giou} \\
 &+ \lambda_5 L_{3dsize} + \lambda_6 L_{angle} + \lambda_7 L_{depth},
\label{2dloss}
\end{aligned}
\end{equation}
where $\lambda_1 \sim \lambda_7$ are the balancing weights. The Laplacian aleatoric uncertainty loss \cite{chen2020monopair} for depth prediction has the following form:
\begin{equation}
	L_{depth} =  \frac{2}{\sigma}||d_{gt} - d_{pre}||_1 + log(\sigma),
	\label{loss_d}
\end{equation}
and the predicted depth value $d_{pre}$ includes three parts:
\begin{equation}
	d_{pre} =  \frac{1}{3}(d_{reg} + d_{geo} + d_{map}),
	\label{dpre_first}
\end{equation}
where $d_{reg}$ is the depth value directly regressed by one detection head, $d_{map}$ get value by the projected 3D center $(x_{3d}, y_{3d})$ and interpolation algorithm on the depth map, and $d_{geo}$ is obtained by the relationship between 2D and 3D sizes:
\begin{equation}
	d_{geo} =  f\frac{h_{3D}}{t+b},
	\label{dpre_second}
\end{equation}
where $f$ is the camera focal length. 

\indent As a result, our composite loss is expressed as:
\begin{equation}
	L = L_{query} + \lambda L_{MSM},
	\label{all loss}
\end{equation}
where $\lambda$ is the balancing weight.

\begin{table*}[ht!]
	\begin{center}
		\begingroup
		\setlength{\tabcolsep}{0.5pt} 
		\renewcommand{\arraystretch}{1.2} 
  \caption{Performance of the ``Car'' category on KITTI test and validation sets, where the first/second best results are highlighted in red/blue fonts, and the improvements are noted in bold fonts.}
		\label{test}
		\begin{tabular}{|m{2.7cm}<{\centering}|m{1.9cm}<{\centering}|m{0.95cm}<{\centering}m{0.95cm}<{\centering}m{0.95cm}<{\centering}|m{0.95cm}<{\centering}m{0.95cm}<{\centering}m{0.95cm}<{\centering}|m{0.95cm}<{\centering}m{0.95cm}<{\centering}m{0.95cm}<{\centering}|m{0.95cm}<{\centering}m{0.95cm}<{\centering}m{0.95cm}<{\centering}|m{0.8cm}<{\centering}|}
			\hline
			\multirow{2}{*}{Method} &
			\multirow{2}{*}{Reference} &
			\multicolumn{3}{c|}{Test, $AP_{3D}$}&
			\multicolumn{3}{c|}{Test, $AP_{BEV}$}&
			\multicolumn{3}{c|}{Val, $AP_{3D}$}&
			\multicolumn{3}{c|}{Val, $AP_{BEV}$}&
			\multirow{2}{*}{\makecell[c]{Time\\(ms)}} \\
			{ } & { }&Easy & Mod. & Hard & Easy & Mod. & Hard &Easy & Mod. & Hard &Easy & Mod. & Hard & {}\\
			\hline\hline
			With extra data:&&&&&&&&&&&&&&\\
			PatchNet \cite{ma2020rethinking} & ECCV2020&15.68 & 11.12 & 10.17 & 22.97 & 16.86 & 14.97 & - & - & - &- & - & - & 400 \\
			D4LCN \cite{ding2020learning} &CVPR2020&16.65 &11.72 &9.51 &22.51 &16.02 &12.55&- & - & - &- & - & - & 200\\
			MonoRUn \cite{chen2021monorun} & CVPR2021&19.65 &12.30 &10.58 &27.94 &17.34 &15.24 &20.02&14.65 &12.61 &- & - & - &70\\
			DDMP-3D \cite{wang2021depth} &CVPR2021& 19.71 &12.78 &9.80 &28.08 &17.89 &13.44& -& -& -&- & - & - &180\\
			CaDDN \cite{reading2021categorical} & CVPR2021&19.17 &13.41 &11.46 &27.94 &18.91 &17.19 &23.57 &16.31 &13.84 &- & - & - &630\\
			AutoShape \cite{liu2021autoshape} &ICCV2021& 22.47 &14.17 &11.36& 30.66& 20.08 &15.59 &20.09& 14.65& 12.07 &- & - & - &-\\
			MonoDTR \cite{huang2022monodtr}  &CVPR2022& 21.99 &15.39 &12.73 &28.59 &20.38 &17.14 &24.52 &18.57 &15.51&33.33 &25.35&21.68&37\\
			OPA-3D \cite{su2023opa} & RAL2023 &24.60&17.05&14.25 &33.54&22.53 &19.22&24.97 &19.40&\textcolor{blue}{16.59}&33.80 &25.51&22.13 &40\\
                MonoNeRD \cite{xu2023mononerd} & ICCV2023&22.75& \textcolor{blue}{17.13}& \textcolor{blue}{15.63}& 31.13& 23.46& 20.97&- & - & - &- & - & -&-\\
			\hline
			Without extra data:&&&&&&&&&&&&&&\\
			MonoFlex \cite{zhang2021objects} & CVPR2021&19.94 &13.89 &12.07 &28.23 &19.75 &16.89 &23.64 &17.51 &14.83 &- & - & - &30\\
			GUPNet \cite{lu2021geometry} & ICCV2021 &20.11 &14.20 &11.77& -& -& - &22.76& 16.46& 13.72 &31.07 &22.94 &19.75&-\\
			DEVIANT \cite{kumar2022deviant} &ECCV2022 &21.88 &14.46 &11.89 &29.65 &20.44 &17.43 &24.63 &16.54 &14.52 &32.60 &23.04 &19.99&40\\
                MonoJSG \cite{lian2022monojsg} & CVPR2022&24.69& 16.14 &13.64 &32.59 &21.26& 18.18 &26.40 &18.30 &15.40&- & - & - &42\\
			DVDET \cite{hu2023aerial} & RAL2023 &23.19&15.44&13.07&32.05&22.15&19.32&- & - & - & - & -& -\\
                MonoDETR \cite{zhang2022monodetr} &ICCV2023&23.65 &15.92 &12.99 &32.08 &21.44 &17.85 &\textcolor{blue}{28.84} &\textcolor{blue}{20.61}&16.38&\textcolor{blue}{37.86}& $\textcolor{blue}{26.95}$&$\textcolor{blue}{22.80}$&\textcolor{red}{20}\\
                PDR \cite{sheng2023pdr}& TCSVT2023&23.69&16.14&13.78&31.76&21.74&18.79&27.65 & 19.44 & 16.24 & 35.59& 25.72& 21.35\\       
                MonoCD \cite{yan2024monocd} &CVPR2024& 25.53 &16.59& 14.53 &33.41 &22.81&19.57& 26.45 &19.37 &16.38& 34.60 &24.96 &21.51 &36\\
                MonoUNI \cite{jinrang2024monouni} &NIPS2024&24.75 &16.73 &13.49 &24.51 &17.18 &14.01&- & - & - &- & - & -&-\\       
                OccupancyM3D \cite{peng2024learning}&CVPR2024& \textcolor{red}{25.55}& 17.02& 14.79 &\textcolor{red}{35.38} &\textcolor{blue}{24.18}& \textcolor{blue}{21.37} &- & - & - &- & - & -&-\\
			S$^3$-MonoDETR & - &\textcolor{blue}{25.54}&\textcolor{red}{17.28} &\textcolor{red}{15.89}&\textcolor{blue}{33.91}&\textcolor{red}{24.59} &\textcolor{red}{21.55} &\textcolor{red}{29.88} &\textcolor{red}{21.51} &\textcolor{red}{18.31} &\textcolor{red}{38.12}&\textcolor{red}{29.72}& \textcolor{red}{25.94}&\textcolor{blue}{26.2}\\
			\hline
			Improvement & vs. Extra \checkmark &$\bm{0.94}$&$\bm{0.15}$&$\bm{0.26}$&$\bm{0.37}$&$\bm{1.13}$& $\bm{0.58}$& $\bm{4.91}$&$\bm{2.11}$&$\bm{1.72}$&$\bm{4.32}$ & $\bm{4.21}$ & $\bm{3.81}$ &$\bm{10.8}$\\
			Improvement & vs. Extra \ding{55} &-&$\bm{0.26}$&$\bm{1.10}$&-&$\bm{0.41}$& $\bm{0.18}$& $\bm{1.04}$&$\bm{0.90}$&$\bm{1.93}$&$\bm{0.26}$ & $\bm{2.77}$&$\bm{3.14}$ &-\\
			\hline
		\end{tabular}
		\endgroup
	\end{center}
\end{table*}
\section{Experiments}
\subsection{Experimental Setup}
\textbf{Dataset:} We validate S$^3$-MonoDETR on the widely recognized KITTI ~\cite{geiger2012we} and Waymo Open~\cite{ettinger2021large} dataset. KITTI consists of $7,481$ training images and $7,518$ testing images. For training and validating our model, we follow the approach proposed in~\cite{chen20153d} and divide the training samples into a sub-training set of $3,712$ images and a validation set of $3,769$ images. We evaluate the detection performance on samples classified into three difficulty levels (easy, moderate, and hard) and assess the precision of the ``Car'', ``Pedestrian'', and ``Cyclist'' categories using average precision (AP) in both 3D space and bird's-eye view (AP$_{3D}$ and AP$_{BEV}$) at $40$ recall positions. The IoU thresholds of these three categories are $0.7$, $0.5$, and $0.5$, respectively.

\indent Waymo Open is the largest autonomous driving dataset which is much more challenging. It contains $798$ training and $202$ validation sequences and generates nearly $160k$ and $40k$ samples, respectively. We follow~\cite{kumar2022deviant} to generate $52,386$ training and $39,848$ validation images from the front camera, where the training images are from every third frame of the training sequences. For evaluation, two difficulty levels ($Level\_1$: points on object $\geq$ 5, and $Level\_2$: points on object $\geq$ 1) are utilized for all the objects, and two IoU thresholds ($0.5, 0.7$) are further used to evaluate four distance ranges ($Overall, 0-30m, 30-50m, 50m-inf$).

\textbf{Implementation Details:}
We employ ResNet-50~\cite{he2016deep} as the feature extraction backbone. The visual encoder and depth encoder are responsible for outputting the visual embedding $\mathbf{F}_V$ and depth embedding $\mathbf{F}_D$, which include three blocks and each block consists of a self-attention layer and a feed-forward neural network (FFN). The depth predictor utilizes two $3 \times 3$ and one $1 \times 1$ convolution layers to output the predicted value and follows \cite{reading2021categorical} to discretize the depth into $k + 1$ bins for faster regression. The decoder consists of a self-attention layer,  the proposed S$^3$-DA layer, and FFN. For S$^3$-DA, we set the number of queries $N$ to $50$, the feature channels $C$ to $256$, and use $8$ heads for the attention layers.
The input images are down-sampled to $1/16$ before being fed into S$^3$-DA, thus in the following parts, $[r, w]$ represents the shape$\&$scale values of objects in the original image after being reduced by $16$ times.

For the preset $[r, w]$ of the ``Car'' category (``Vehicle'' in Waymo Open), we select six combinations: $\{[1, 1], [1, 2], [1, 4], [1, 6], [0.5, 4], [0.5, 8]\}$, and additionally add $\{[2, 2], [3, 2], [2, 4]\}$ for multi-category joint training.
As for the single process training for the ``Pedestrian'' and ``Cyclist'', we use $\{[2, 2], [2, 4], [3, 2]\}$ and $\{[1, 2], [1, 4], [2, 2]\}$, respectively.

During model training, the balancing weights $\lambda_1 \sim \lambda_7$ in Equation~\ref{2dloss} are set to $2, 5, 10, 2, 1, 1, 1$, and $\lambda$ in Equation~\ref{all loss} is set to $0.5$, whereas $W_1$ and $W_2$ in Equation~\ref{category_label_2} are set to $2$ and $1$. We use a single RTX A6000 GPU and adopt the Adam optimizer with a weight decay of $10^{-4}$ as the training strategy. When training on KITTI, we conduct $180$ epochs with a batch size of $16$ and an initial learning rate of $2 \times 10^{-4}$. We train our model on Waymo Open for $40$ epochs, the batch size and initial learning rate are set to $40$ and $5 \times 10^{-4}$, respectively.

\subsection{Performance Comparison}
Table~\ref{test} shows the performance comparison results of the ``Car'' category on the KITTI test and validation sets.
From the table, we can observe that S$^3$-MonoDETR achieves the best accuracy on moderate and hard objects, especially for the hard difficulty, our method achieves an improvement of up to $3.81\%$.
This is because hard samples usually have smaller sizes, and thus the feature extraction for query points by the previous algorithms is prone to introduce noise from the background or surrounding objects, resulting in detection errors.
Comparatively, S$^3$-DA could achieve superior shape$\&$scale perception ability for query points in a supervised manner, thus enabling the model to attain significant performance improvement. 

\begin{table}[t!]
	\begin{center}
		\begingroup
		\setlength{\tabcolsep}{0.5pt} 
		\renewcommand{\arraystretch}{1.2} 
		\caption{Performance of the ``Pedestrian'' and ``Cyclist'' categories on KITTI test set, where ``v.$\checkmark$'' and ``v.\ding{55}'' represent the accuracy improvement compared to the methods with/without extra data. The first/second best results are highlighted in red/blue fonts, and the improvements are noted in bold fonts.}
        \label{test2}
		\begin{tabular}{|m{2.7cm}<{\centering}|m{0.95cm}<{\centering}m{0.95cm}<{\centering}m{0.95cm}<{\centering}|m{0.95cm}<{\centering}m{0.95cm}<{\centering}m{0.95cm}<{\centering}|}
			\hline
			\multirow{2}{*}{Method} &
			\multicolumn{3}{c|}{Pedestrian, $AP_{3D}$}&
			\multicolumn{3}{c|}{Cyclist, $AP_{3D}$} \\
			{ } & Easy & Mod. & Hard & Easy & Mod. & Hard \\
			\hline\hline
			With extra data:&&&&&&\\
            D4LCN \cite{ding2020learning}  &4.55 &3.42 &2.83 &2.45 &1.67 &1.36 \\
            DDMP-3D \cite{wang2021depth} &4.93 &3.55 &3.01 &4.18 &2.50& 2.32\\
			MonoPSR \cite{ku2019monocular} & 6.12 &4.00 &3.30& \textcolor{red}{8.70} &\textcolor{blue}{4.74}& \textcolor{blue}{3.68} \\
            CADNN \cite{reading2021categorical} &\textcolor{blue}{12.87} &8.14 &6.76 &7.00& 3.41 &3.30\\
			\hline
			Without extra data:&&&&&&\\
            MonoGeo \cite{zhang2021learning} &8.00 &5.63 &4.71 &4.73 &2.93 &2.58\\
			MonoFlex \cite{zhang2021objects} & 9.43 &6.31 &5.26& 4.17 &2.35& 2.04 \\
            MonoDLE \cite{ma2021delving} & 9.64 &6.55 &5.44 &4.59 &2.66 &2.45 \\
            MonoPair \cite{chen2020monopair} &10.02 &6.68 &5.53& 3.79 &2.12 &1.83\\
			MonoDETR \cite{zhang2022monodetr}&12.54& 7.89& 6.65& 7.33& 4.18& 2.92 \\
			DEVIANT \cite{kumar2022deviant} & \textcolor{red}{13.43} &\textcolor{blue}{8.65} &\textcolor{blue}{7.69} &5.05 &3.13 &2.59\\
   S$^3$-MonoDETR &12.62&$\textcolor{red}{9.85}$ &$\textcolor{red}{8.61}$ &$\textcolor{blue}{7.68}$ &$\textcolor{red}{5.89}$ & $\textcolor{red}{4.12}$ \\
			\hline
			Improvement v.\checkmark &-& $\bm{1.71}$&$\bm{1.85}$&-& $\bm{1.15}$ & $\bm{0.44}$ \\
			Improvement v.\ding{55} &-& $\bm{1.20}$ & $\bm{0.92}$ &$\bm{0.35}$& $\bm{1.71}$ & $\bm{1.20}$ \\
			\hline
		\end{tabular}
		\endgroup
	\end{center}
\end{table}

\indent Table \ref{test2} further demonstrates the performance on ``Pedestrian'' and ``Cyclist'' categories.
Obviously, these two categories are much more difficult than ``Car'' due to their smaller size and non-rigid body.
Thanks to the effective S$^3$-DA, our method outperforms all the without-extra-data methods on three difficulty levels, especially for the moderate and hard objects, where we achieve around a $2\%$ improvement. The results in Table~\ref{test2} verify the outstanding generality and expansibility of our model, which only requires readily obtainable prior knowledge of different categories of shapes and scales.
With this, our method can effortlessly perform accurate detection on objects with various appearances. \\
\indent Despite the addition of finely designed components, our method still has strong competitiveness in terms of speed compared to all existing methods as shown in Table~\ref{test}, only $5ms$ slower than MonoDETR per image. 
While S$^3$-DA brings extra computation costs, our proposed model avoids the usage of the depth cross-attention layer in MonoDETR, which results in a similar inference speed under the same hardware conditions. \\
\indent As it can be observed, the large improvements of our method appear mostly in the ``Moderate'' and ``Hard'' subsets.
This is due to the difficulty classification criteria of the KITTI dataset, where the ``Easy'' objects are mostly close and their degree of occlusion by other objects is $0$.
Thus, the ``Easy'' objects usually have high-quality image features and are slightly suffered by the surrounding objects, which results in the proposed S$^3$-DA not being able to present an obvious effect as it does in the ``Moderate'' and ``Hard'' subsets where our method shines with consistent and significant gains.

\begin{table*}[ht]
	\begin{center}
		\begingroup
		\setlength{\tabcolsep}{0.5pt} 
		\renewcommand{\arraystretch}{1.2} 
  \caption{Performance comparison of the ``Car'', ''Pedestrian'', and ``Cyclist'' categories joint training on KITTI validation set, where the first/second best results are highlighted in red/blue fonts, and the improvements are noted in bold fonts.}
  \label{multi2}
		\begin{tabular}{|m{2.7cm}<{\centering}||m{0.95cm}<{\centering}m{0.95cm}<{\centering}m{0.95cm}<{\centering}m{0.95cm}<{\centering}||m{0.95cm}<{\centering}m{0.95cm}<{\centering}m{0.95cm}<{\centering}m{0.95cm}<{\centering}||m{0.95cm}<{\centering}m{0.95cm}<{\centering}m{0.95cm}<{\centering}m{0.95cm}<{\centering}||m{1.8cm}<{\centering}|}
			\hline
			\multirow{2}{*}{Method} &
			\multicolumn{4}{|c|}{Car}&
			\multicolumn{4}{|c|}{Pedestrian}&
			\multicolumn{4}{|c|}{Cyclist}&
            \multirow{2}{*}{3class-mAP} \\
			{ } & Easy & Mod. & Hard &mAP& Easy & Mod. & Hard &mAP&Easy & Mod. & Hard &mAP& { } \\
			\hline\hline
            MonoFlex \cite{zhang2021objects} & 24.22 &17.34 &15.14 &18.90 & 6.80 &4.89  &4.09 &5.26 &7.90 &4.05 &3.85&5.27&9.81\\
			GUPNet \cite{lu2021geometry}     & 23.19 &16.23 &13.57 &17.66 &11.29 &7.05  &6.36 &8.23 &9.49 &5.01 &4.14&$\textcolor{blue}{6.21}$&10.70\\
            MonoDTR \cite{huang2022monodtr}  & 23.83 &17.24 &14.30 &18.46 &14.13 &10.61 &8.71 &$\textcolor{red}{11.15}$&6.88 &3.99 &4.00&4.96&$\textcolor{blue}{11.52}$\\
			MonoDETR \cite{zhang2022monodetr}& 24.62 &18.11 &14.91 &$\textcolor{blue}{19.21}$ &7.48  &5.63  &4.56 &5.89 &7.45 &3.90 &3.80&5.05&10.05\\
			S$^3$-MonoDETR                   & 26.50 &18.87 &15.78 &$\textcolor{red}{20.38}$ &11.87 &9.50  &7.88 &$\textcolor{blue}{9.75}$ &10.75&5.99 &5.85&$\textcolor{red}{7.53}$&$\textcolor{red}{12.55}$\\
			\hline
			Improvement &with& second               &        &$\bm{1.17}$&    &      &     &-&   & & &$\bm{1.32}$&$\bm{1.03}$\\
            Improvement &with& baseline             &        &$\bm{1.17}$&    & &&$\bm{3.86}$&   & & &$\bm{2.48}$&$\bm{2.50}$\\
			\hline
		\end{tabular}
		\endgroup
	\end{center}
\end{table*}

\begin{table*}[t!]
	\begin{center}
    \begingroup
    \setlength{\tabcolsep}{0.5pt} 
    \renewcommand{\arraystretch}{1.2} 
	\caption{``Vehicle'' performance on the Waymo Open Val set between different models. We use average precision on 3D view ($AP_{3D}$) ($LEVEL\_1$ and $LEVEL\_2$, IoU \textgreater 0.5 and IoU \textgreater 0.7) according to four object distance intervals.}
 \label{waymo1}
\begin{tabular}{|m{2.5cm}<{\centering}|m{1.3cm}<{\centering}|m{0.8cm}<{\centering}m{0.8cm}<{\centering} m{0.9cm}<{\centering}m{0.94cm}<{\centering}|m{0.8cm}<{\centering}m{0.8cm}<{\centering} m{0.9cm}<{\centering}m{0.94cm}<{\centering}|m{0.8cm}<{\centering}m{0.8cm}<{\centering} m{0.9cm}<{\centering}m{0.94cm}<{\centering}|m{0.8cm}<{\centering}m{0.8cm}<{\centering} m{0.9cm}<{\centering}m{0.94cm}<{\centering}|}
\hline
\multirow{2}{*}{Method} &
\multirow{2}{*}{Reference} &
\multicolumn{4}{c|}{$LEVEL\_1$(IoU \textgreater 0.5)}&
\multicolumn{4}{c|}{$LEVEL\_2$(IoU \textgreater 0.5)}&
\multicolumn{4}{c|}{$LEVEL\_1$(IoU \textgreater 0.7)}&
\multicolumn{4}{c|}{$LEVEL\_2$(IoU \textgreater 0.7)}\\
{ } &{ }&Overall &0-30m &30-50m & 50m-Inf&Overall &0-30m &30-50m & 50m-Inf&Overall &0-30m &30-50m & 50m-Inf&Overall &0-30m &30-50m & 50m-Inf \\
\hline\hline
With extra data:&&&&& &&&& &&&& &&&&\\
PatchNet~\cite{ma2020rethinking} &ECCV 20 &2.92 &10.03 &1.09 &0.23 &2.42 &10.01 &1.07 &0.22& 0.39 &1.67 &0.13 &0.03 &0.38 &1.67 &0.13 &0.03\\
PCT~\cite{wang2021progressive} &NIPS 21 & 4.20 &14.70 &1.78 &0.39 &4.03 &14.67 &1.74 &0.36& 0.89 &3.18 &0.27 &0.07&0.66 &3.18 &0.27 &0.07 \\
\hline
Without extra data:&&&&& &&&& &&&& &&&&\\
GUPNet~\cite{lu2021geometry} &ICCV 21& 10.02 &24.78 &4.84 &0.22 &9.39 &24.69 &4.67 &0.19& 2.28 &6.15 &0.81 &0.03& 2.14 &6.13 &0.78& 0.02\\
MonoJSG~\cite{lian2022monojsg} &CVPR 22& 5.65 &20.86 &3.91 &$\textcolor{blue}{0.97}$ &5.34 &20.79 &3.79 &$\textcolor{blue}{0.85}$&0.97 &4.65 &0.55 &$\textcolor{blue}{0.10}$ &0.91 &4.64 &0.55 &$\textcolor{blue}{0.09}$\\
DEVIANT~\cite{kumar2022deviant} &ECCV 22 &10.98 &26.85 &$\textcolor{blue}{5.13}$ &0.18 &10.29 &26.75 &$\textcolor{blue}{4.95}$ &0.16&2.69 &6.95 &$\textcolor{blue}{0.99}$ &0.02 &2.52 &6.93 &$\textcolor{blue}{0.95}$ &0.02\\
MonoRCNN++~\cite{shi2023multivariate} &WACV 23&$\textcolor{blue}{11.37}$ &$\textcolor{red}{27.95}$ &4.07 &0.42 &$\textcolor{blue}{10.79}$ &$\textcolor{red}{27.88}$ &3.98& 0.39&$\textcolor{blue}{4.28}$ &$\textcolor{blue}{9.84}$ &0.91 &0.09 &$\textcolor{blue}{4.05}$ &$\textcolor{red}{9.81}$ &0.89 &0.08\\
S$^3$-MonoDETR& - &$\textcolor{red}{11.65}$&$\textcolor{blue}{27.82}$ &$\textcolor{red}{5.15}$ &$\textcolor{red}{0.98}$& $\textcolor{red}{11.12}$&$\textcolor{blue}{27.71}$ &$\textcolor{red}{5.23}$&$\textcolor{red}{0.88}$&$\textcolor{red}{4.48}$ &$\textcolor{red}{9.96}$&$\textcolor{red}{1.12}$ &$\textcolor{red}{0.17}$&$\textcolor{red}{4.13}$ &$\textcolor{blue}{9.61}$&$\textcolor{red}{1.10}$&$\textcolor{red}{0.11}$ \\
\hline
\end{tabular}
\endgroup
\end{center}
\end{table*}

\begin{table*}[t!]
	\begin{center}
    \begingroup
    \setlength{\tabcolsep}{0.5pt} 
    \renewcommand{\arraystretch}{1.2} 
	\caption{``Pedestrian'' and ``Cyclist'' performance on the Waymo Open Val set between different models. We use $AP_{3D}$ ($LEVEL\_1$ and $LEVEL\_2$, IoU \textgreater 0.5) according to four object distance intervals.}
 \label{waymo2}
\begin{tabular}{|m{2.5cm}<{\centering}|m{1.3cm}<{\centering}|m{0.8cm}<{\centering}m{0.8cm}<{\centering} m{0.9cm}<{\centering}m{0.94cm}<{\centering}|m{0.8cm}<{\centering}m{0.8cm}<{\centering} m{0.9cm}<{\centering}m{0.94cm}<{\centering}|m{0.8cm}<{\centering}m{0.8cm}<{\centering} m{0.9cm}<{\centering}m{0.94cm}<{\centering}|m{0.8cm}<{\centering}m{0.8cm}<{\centering} m{0.9cm}<{\centering}m{0.94cm}<{\centering}|}
\hline
\multirow{2}{*}{Method} &
\multirow{2}{*}{Reference} &
\multicolumn{4}{c|}{Pedestrian($LEVEL\_1$)}&
\multicolumn{4}{c|}{Pedestrian($LEVEL\_2$)}&
\multicolumn{4}{c|}{Cyclist($LEVEL\_1$)}&
\multicolumn{4}{c|}{Cyclist($LEVEL\_2$)}\\
{ } &{ }&Overall &0-30m &30-50m & 50m-Inf&Overall &0-30m &30-50m & 50m-Inf&Overall &0-30m &30-50m & 50m-Inf&Overall &0-30m &30-50m & 50m-Inf \\
\hline\hline
GUPNet~\cite{lu2021geometry} &ICCV 21& \textcolor{blue}{4.28} &\textcolor{blue}{11.84}  &1.82  &0.01 &\textcolor{blue}{3.89} &\textcolor{blue}{11.66} &1.70 &0.01& 2.96  &8.81  &0.05  &0.00& 2.85  &8.76  &0.05  &0.00\\
DEVIANT~\cite{kumar2022deviant} &ECCV 22 &4.03 &10.15 &\textcolor{blue}{2.42} &0.01 &3.67 &10.00 &\textcolor{blue}{2.26} &0.01 &\textcolor{blue}{5.12} &\textcolor{blue}{13.14} &\textcolor{blue}{1.74} &0.00 &\textcolor{blue}{4.93} &\textcolor{red}{13.08} &\textcolor{blue}{1.65} &0.00\\
S$^3$-MonoDETR& - &$\textcolor{red}{4.42}$&$\textcolor{red}{12.12}$ &$\textcolor{red}{2.68}$ &0.01& $\textcolor{red}{4.01}$&$\textcolor{red}{11.89}$ &$\textcolor{red}{2.69}$&0.01&$\textcolor{red}{5.32}$ &$\textcolor{red}{13.52}$&$\textcolor{red}{1.90}$ &0.00&$\textcolor{red}{4.95}$ &$\textcolor{blue}{13.06}$&$\textcolor{red}{1.85}$&0.00 \\
\hline
\end{tabular}
\endgroup
\end{center}
\end{table*}

\begin{figure}[!t]
	\centering
	\includegraphics[scale=0.4]{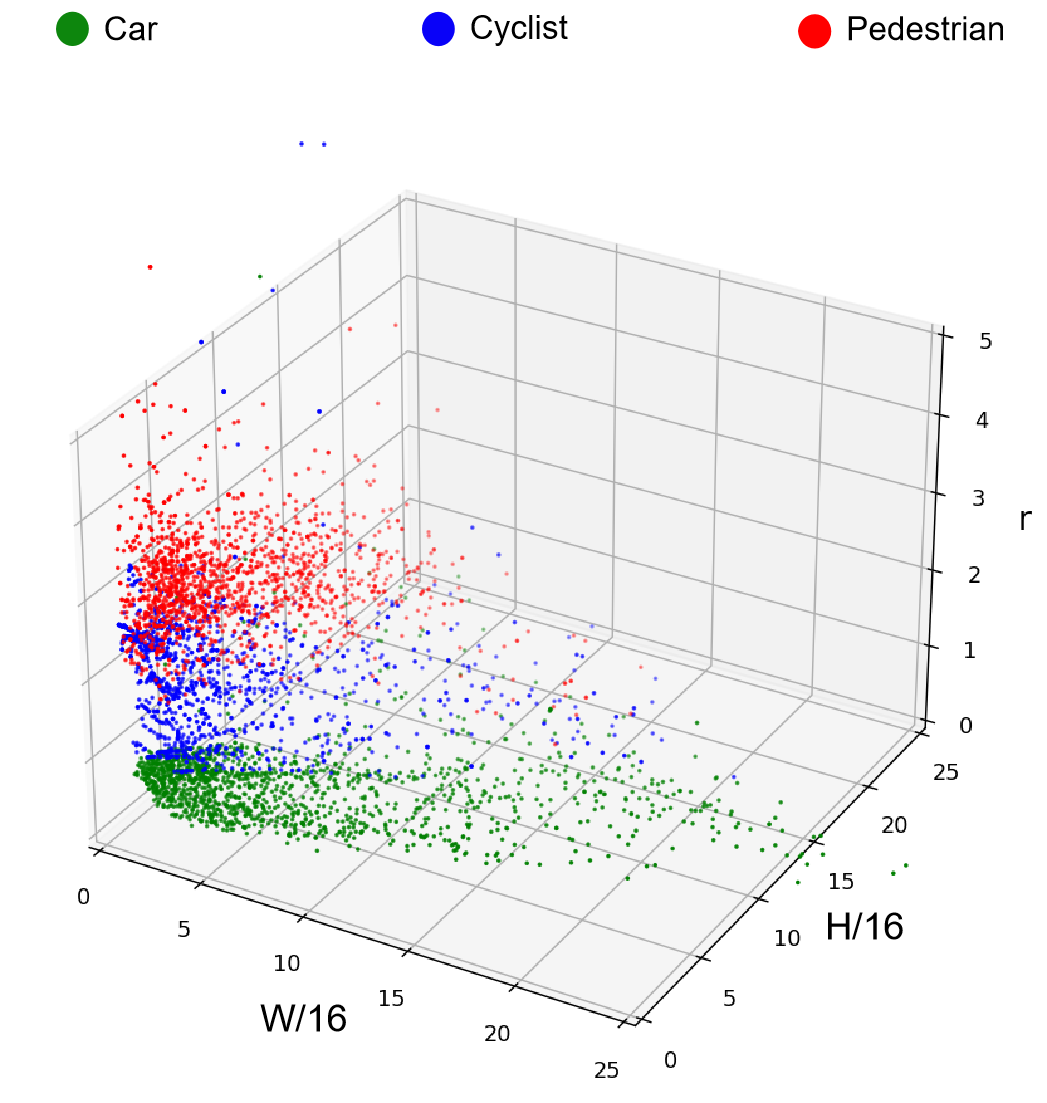}
	\caption{The shape and scale distribution of ``Car'', ``Pedestrian'', and ``Cyclist'' in the camera view.}
	\label{3class}
\end{figure}
\subsection{Evaluation of Multi-category Joint Training}
\indent As shown in Fig.~\ref{3class}, in the camera view, there are significant differences in the distribution space of shape (\textit{i.e.}, the aspect ratio $r$) and scale (\textit{i.e.}, the width $w$ and height $h$) among ``Car'', ``Pedestrian'', and ``Cyclist''.
Therefore, as our method is dedicated to enhancing query points by the valuable information of object appearance, it also has a strong capability of multi-category joint training in a single process as long as we preset appropriate shape$\&$scale for different categories of objects, which is a weakness in the baseline MonoDETR.
Specifically, on the basis of the preset shape$\&$scale for the ``Car'' category, we extra add $\{[2, 2], [3, 2], [2, 4]\}$ for ``Pedestrian'' and ``Cyclist'' categories.\\
\indent Table~\ref{multi2} illustrates the performance comparison between several state-of-the-art methods, for which we train three categories of objects simultaneously in a single process.
It can be easily discovered that most methods have a ``category imbalance'' issue, such as MonoDTR achieving the highest mAP on ``Pedestrian'', yet, the lowest mAP on ``Cyclist''.
Furthermore, our baseline MonoDETR achieves very high performance on ``Car'' but poor performance on the other categories, reflecting the limitation of the unsupervised attention mechanism for detecting multi-category objects.
Under comparison, our method achieves superior performance in all categories thanks to the supervised shape$\&$scale-perceptive mechanism, which could impose targeted feature augmentation for all categories of objects, thus effectively solving the network learning challenge caused by the significant differences in the appearance of different objects in a monocular image.

\indent We further evaluate our method on the Waymo Open Val set. Table~\ref{waymo1} lists the ``Vehicle'' detection performance and Table~\ref{waymo2} shows the results on the ``Pedestrian'', and ``Cyclist''. Note that the results of three categories are obtained from a single process training. For all the categories evaluated on two difficulty levels, our S$^3$-MonoDETR yields superior breakthroughs against all the other monocular 3D detectors in terms of ``Overall'' perspective, which proves the effectiveness of our proposed S$^3$-DA layer and MSM loss. On the distance $30m \sim inf$, our S$^3$-MonoDETR also exceeds all the other methods, which is consistent with the design motivation of our proposed shape$\&$scale-perceptive mechanism, generating higher-quality query features thus obviously improving the accuracy for those hard and distant objects. Moreover, from Table~\ref{waymo2}, it can be observed that GUPNet and DEVIANT achieve better performance on ``Pedestrians'' and ``Cyclist'', respectively. However, our S$^3$-MonoDETR achieves superior performance in all categories, which further proves the robustness of our proposed method.

\subsection{Evaluation of S$^3$-DA}
In this part, all the experiments are conducted on the ``Car'' category on the KITTI validation set, and we report the AP$_{3D}$ results.
We first try different shape$\&$scale settings in S$^3$-DA to observe the performance trend and then evaluate the quality of the generated query points by S$^3$-DA. Finally, we evaluate the effectiveness of the MCM loss.\\
\begin{table}[ht]
	\begin{center}
		\renewcommand{\arraystretch}{1.2}
  \caption{Performance change by using different shape$\&$scale settings for the ``Car'' category.}
		\setlength{\tabcolsep}{2.5pt}{
			\begin{tabular}{|m{2cm}<{\centering}|m{3cm}<{\centering}m{1cm}<{\centering}m{1cm}<{\centering}|m{1cm}<{\centering}|}
				\hline
            \multicolumn{2}{c|}{[r, w]}&Easy&Mod.&Hard \\
			common & extra &{}&{}&{}\\
				\hline\hline
                -&- & 27.07 & 19.31 & 16.23\\
                \hline
				\multirow{5}{*}{\makecell[c]{$[1, 1], [1, 2],$ \\ $[1, 4], [1, 6]$}} 
                &-                      & 27.92 & 19.68 & 17.41 \\
                &[1, 8]                & 28.02 & 19.34 & 17.20 \\
			&[0.5, 4]                & 28.89 & 19.93 & 17.86 \\
                &[0.5, 4], [0.5, 8]        & $\bm{29.88}$ & $\bm{21.51}$ & $\bm{18.31}$\\
                &[0.5, 4], [0.5, 8], [0.5, 12]& 29.69& 21.03& 17.89\\
				\hline
		\end{tabular}}
		\label{ab2}
	\end{center}
\end{table}

\textbf{Shape$\&$Scale Settings in S$^3$-DA}: As shown in Fig.~\ref{3class}, over $90\%$ objects are reduced within $1$ to $14$ pixels in scale. For the ``Car'' category, its aspect ratio $r$ is concentrated between $0.3$ and $1$. Based on this, as shown in Table~\ref{ab2}, we first fix the use of the four most common $[r, w]$, \textit{i.e.}, $[1, 1], [1, 2],[1, 4], [1, 6]$, and this combination achieves performance that exceeds the baseline by $1.18\%$ in the hard level, which is because many queries located on small objects receive the crucial feature enhancement.
Then, we continuously add $[1, 8]$ but see overall performance degradation, which is due to that the large-sized cars in the image are very close to the camera, which are mostly on both sides of the camera and presented at an oblique angle with small $r$ in the image.
So, for larger scales ($w \textgreater 6$), we only set $r=0.5$ to pair with them.
When we add the $[r, w]$ of $[0.5, 4]$ and $[0.5, 8]$, the performance continues to increase.
However, the performance begins to degrade as $[0.5, 12]$ is added, which is attributed to that this combination of $[r, w]$ exceeds most objects and would interrupt the shape$\&$scale estimation on object queries.\\
\indent This experiment proves that we do not need to preset a large number of shapes and sizes with wide coverage. A rough estimation combined with the following adaptive learning layers can effectively enhance the query features with shape$\&$scale-perception.

\textbf{Quality Evaluation of Key Points}:
We evaluate the quality of query points using three approaches: MonoDETR, S$^3$-MonoDETR/MCM (S$^3$-MonoDETR without using MCM loss), and the complete S$^3$-MonoDETR.
Table~\ref{ER} presents the comparison results, which are assessed using two quantitative criteria: Position Precision and Weighted Position Precision. Position precision represents the proportion of key points falling inside objects, while Weighted Position Precision extra assigns the corresponding attention weight to each key point for calculating position precision.
Thanks to the incorporation of S$^3$-DA, S$^3$-MonoDETR demonstrates accurate estimation of shape$\&$scale for queries, resulting in better position precision of key points compared to MonoDETR.
However, when the MCM loss is not utilized, there is a significant drop in performance, highlighting the effectiveness of the proposed MCM loss in supervising the shape$\&$scale prediction.
Additionally, we observe that the improvement in Weighted Position Precision is greater than that in Position Precision, indicating that S$^3$-MonoDETR assigns higher attention weights to key points located inside objects.

\begin{table}[ht]
 \begin{center}
  \renewcommand{\arraystretch}{1.2}
  \caption{The prediction accuracy of key points by MonoDETR, our method without MCM loss, and our complete method.}
  \label{ER}
  \setlength{\tabcolsep}{2.5pt}{
   \begin{tabular}{|m{3.5cm}<{\centering}|m{1.5cm}<{\centering}m{3cm}<{\centering}|}
    \hline
    Method              &  Position Precision & Weighted Position Precision \\
    \hline\hline
    MonoDETR            & 70.17 $\%$  & 74.61$\%$  \\
    S$^3$-MonoDETR/MCM   & 74.18 $\%$ & 85.98 $\%$  \\
    S$^3$-MonoDETR                  & $\bm{78.12\%}$ & $\bm{94.23\%}$  \\
    \hline
  \end{tabular}}
 \end{center}
\end{table}

\textbf{Evaluation of the MCM Loss:}
In this experiment, we investigate with different weights $\lambda$ of MSM loss in Equation~\ref{all loss}.
Table~\ref{ab1} illustrates the performance trend across different values of $\lambda$.
As $\lambda$ increases from $0$ to $0.1$, the performance exhibits a significant improvement of approximately $2.5\%$ across all samples. This improvement can be attributed to the utilization of the MSM loss, which effectively supervises the prediction of shape and scale for query points, thereby providing higher-quality query features for the following detection heads.
However, excessive weights on MSM loss will lead to performance degradation.
This is because an overly dominant MSM loss can overshadow the contribution of $L_{2D}$ and $L_{3D}$ detection losses, introducing prediction errors.
Therefore, striking the right balance for the weighting of the MSM loss is crucial to reach reliable object detection.

\begin{table}[ht]
	\begin{center}
		\renewcommand{\arraystretch}{1.2}
  \caption{Performance trend with respect to different values of $\lambda$.}
  \label{ab1}
		\setlength{\tabcolsep}{2.5pt}{
			\begin{tabular}{|m{1cm}<{\centering}|m{1.5cm}<{\centering}m{1.5cm}<{\centering}m{1.5cm}<{\centering}|}
				\hline
				$\lambda$ & Easy & Mod. & Hard \\
				\hline\hline
				0   & 27.12 & 19.01 & 16.15 \\
				0.1 & $\bm{29.88}$ & $\bm{21.51}$ & $\bm{18.31}$\\
                    0.2 & 29.30& 21.13 & 17.87\\
				0.3 & 28.64& 20.45 & 17.12  \\
				0.4 & 27.56& 19.48 & 16.30 \\
				\hline
		\end{tabular}}
	\end{center}
\end{table}

\begin{figure*}[t!]
	\centering
	\begin{minipage}{0.495\linewidth}
		\centering
		\includegraphics[width=1\linewidth]{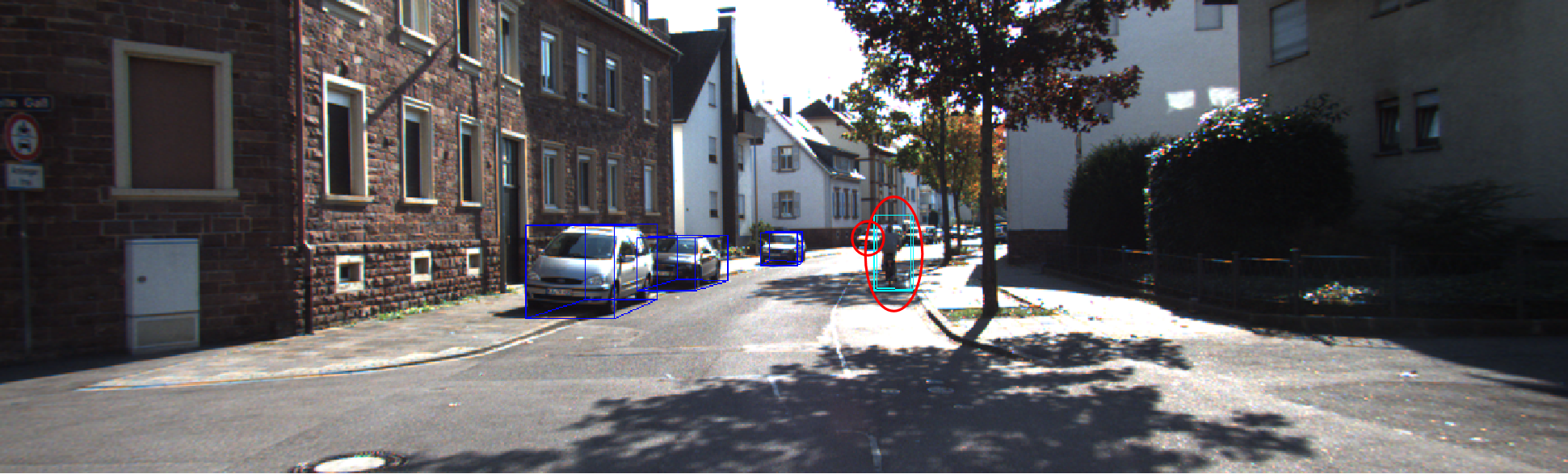}\vspace{4pt}
	\end{minipage}
	\begin{minipage}{0.495\linewidth}
		\centering
		\includegraphics[width=1\linewidth]{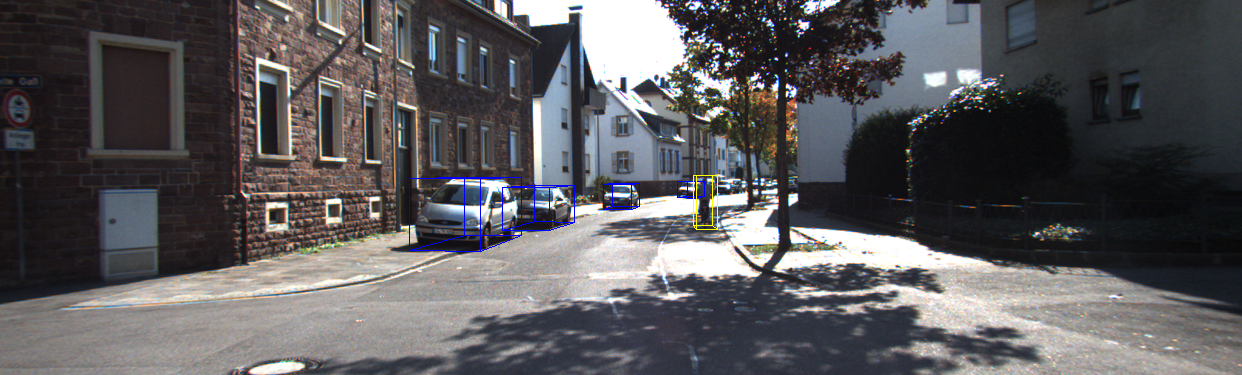}\vspace{4pt}
	\end{minipage}
	\begin{minipage}{0.495\linewidth}
		\centering
		\includegraphics[width=1\linewidth]{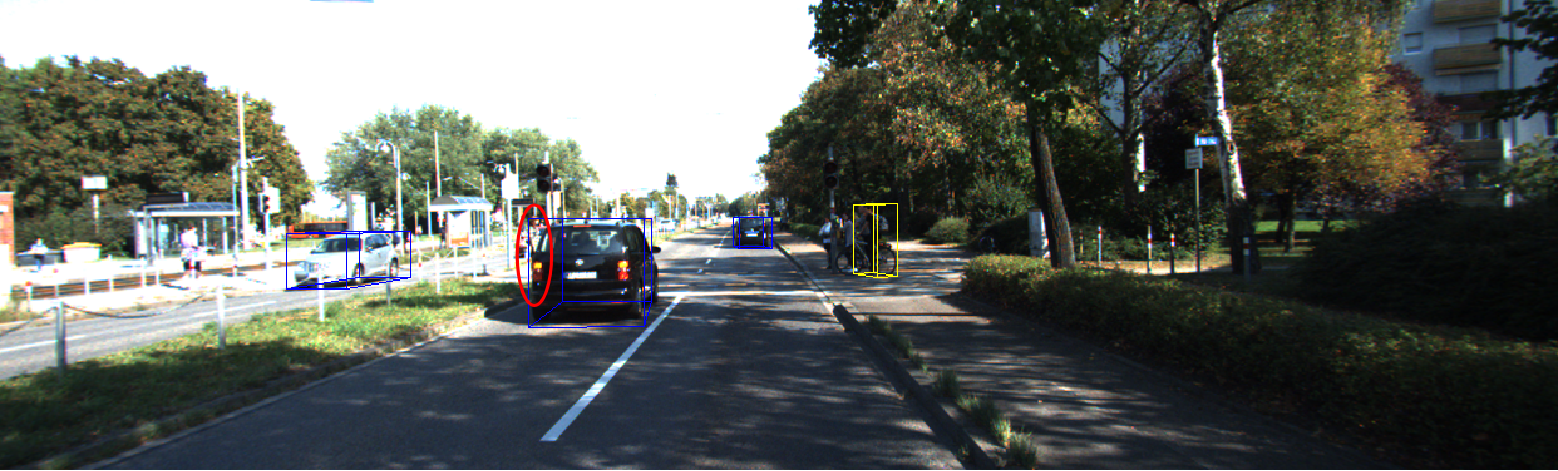}\vspace{4pt}
	\end{minipage}
	\begin{minipage}{0.495\linewidth}
		\centering
		\includegraphics[width=1\linewidth]{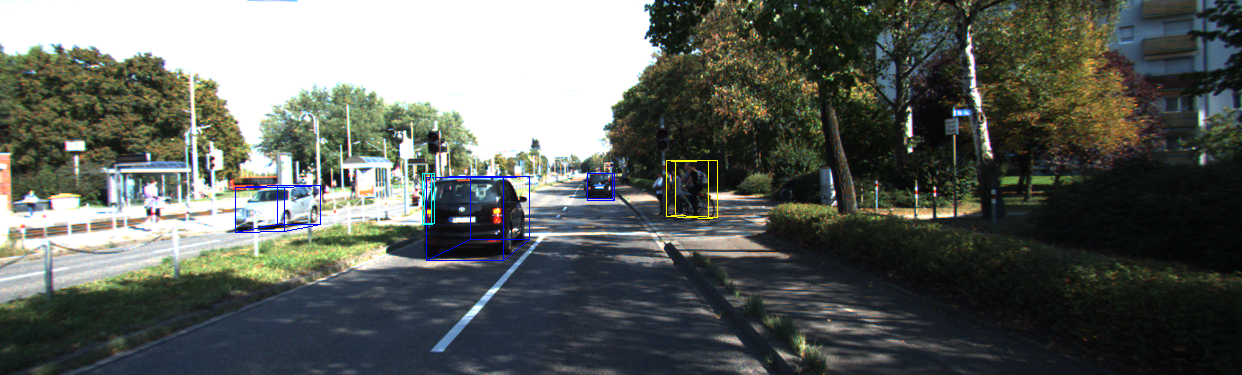}\vspace{4pt}
	\end{minipage}
	\begin{minipage}{0.495\linewidth}
		\centering
		\includegraphics[width=1\linewidth]{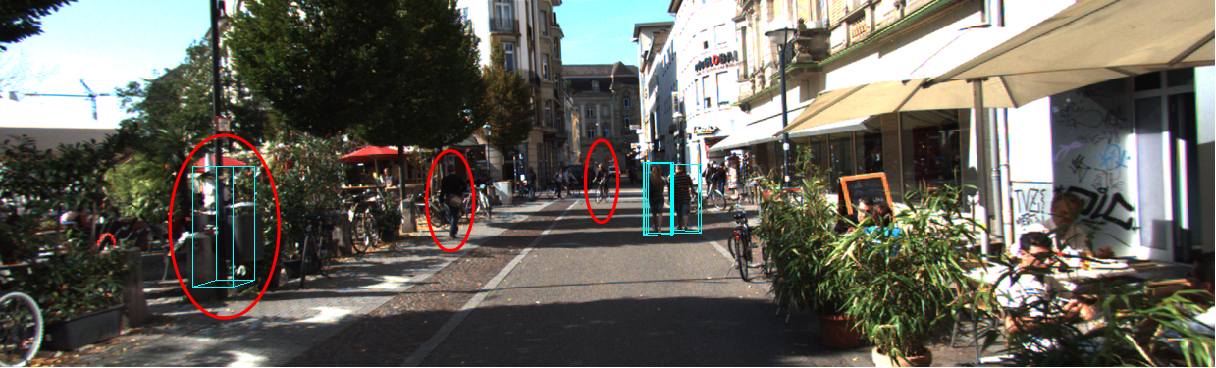}\vspace{4pt}
	\end{minipage}
	\begin{minipage}{0.495\linewidth}
		\centering
		\includegraphics[width=1\linewidth]{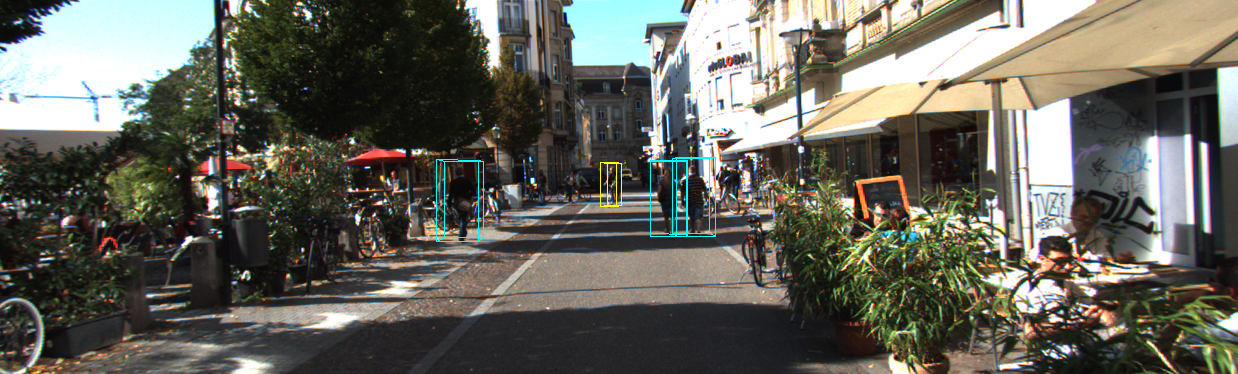}\vspace{4pt}
	\end{minipage}
	\caption{Four representative examples to visualize the detection results between MonoDETR (left) and S$^3$-MonoDETR (right), where the red circles indicate the missing or false-detected objects. ``Car'' (dark blue), ``Pedestrian'' (light blue), ``Cyclist'' (yellow).}
	\label{duibi}
\end{figure*}

\subsection{Qualitative Results}
Fig.~\ref{duibi} presents three visual examples to show the detection results obtained by multi-category joint training of MonoDETR and S$^3$-MonoDETR.
We observe that MonoDETR misses or falsely detects several small or occluded objects, which is due to that the unsupervised attention mechanism in it can not adaptively extract high-quality visual features for objects with huge appearance differences, and is particularly fatal for those hard samples.
In contrast, S$^3$-MonoDETR effectively estimates the object's shape and scale, allowing it to extract relevant local features from within the object.
Consequently, S$^3$-MonoDETR produces robust query features for objects with various appearances in an image, enabling accurate detection even for challenging samples.

\section{Conclusion}
We propose S$^3$-MonoDETR, which is dedicated to improving the query quality in transformer-based monocular 3D object detection for the first time. Unlike existing transformer-based methods, S$^3$-MonoDETR introduces a Supervised Shape$\&$Scale-perceptive Deformable Attention (S$^3$-DA), which could estimate the shape and scale of a query to capture its shape$\&$scale-perceptive receptive field with an accurate scope. Benefiting from this, the query could aggregate high-quality local features, resulting in more precise 3D attribute prediction as well as enabling the model to detect multi-category objects with significant appearance differences in a single training process. Moreover, S$^3$-DA proposes a query-level fine-grained feature fusion mechanism and an MCM loss to achieve the accurate estimation of query shape$\&$scale without requiring additional labeling cost. Extensive experiments conducted on KITTI and Waymo Open datasets demonstrate the effectiveness and near real-time inference capability of our method, showcasing its high practical value for autonomous driving applications. In the future, we intend to extend our approach to 4D-millimeter-radar-based object detection for self-driving scene understanding.

\bibliographystyle{IEEEtran}
\bibliography{egbib}

\end{document}